# Statistical Modelling in Continuous Speech Recognition (CSR)

**Steve Young**
Cambridge University Engineering Dept.
Trumpington Street, Cambridge
England CB2 1PZ
sjy@eng.cam.ac.uk

## Abstract

Automatic continuous speech recognition (CSR) is sufficiently mature that a variety of real world applications are now possible including large vocabulary transcription and interactive spoken dialogues. This paper reviews the evolution of the statistical modelling techniques which underlie current-day systems, specifically hidden Markov models (HMMs) and N-grams. Starting from a description of the speech signal and its parameterisation, the various modelling assumptions and their consequences are discussed. It then describes various techniques by which the effects of these assumptions can be mitigated. Despite the progress that has been made, the limitations of current modelling techniques are still evident. The paper therefore concludes with a brief review of some of the more fundamental modelling work now in progress.

## 1 INTRODUCTION

The aim of this paper is two-fold. The first aim is to set out the basic framework for the statistical approach to speech recognition, identify its limitations and describe how they can be mitigated in practical implementations. The second aim is to indicate some of the directions in which new models might evolve.

The foundations of modern speech recognition technology were laid by Fred Jelinek and his team at IBM in the 1970's[1]. Reflecting the computational power of the time, initial development in the 1980's focussed on whole word small vocabulary applications[2]. In the early 90's attention switched to continuous speaker-independent recognition. Starting with the artificial 1000 word *Resource Management* task, the technology developed rapidly and by the mid-1990's, reasonable accuracy was being achieved for unrestricted dictation.

Much of this development was driven by a series of DARPA and NSA programmes[3] which set ever more challenging tasks culminating most recently in systems for transcribing broadcast news programmes[4] and for transcribing spontaneous telephone conversations[5].

Although the basic framework for CSR has not changed significantly in the last ten years, the detailed modelling techniques developed within this framework have evolved to a state of considerable sophistication. The result has been steady and significant progress. The paper continues in section 2 with a brief review of this framework and its limitations. Then in section 3, the major areas of refinement employed by today's state-of-the-art systems are discussed.

Despite the progress that has been made, the limitations of current modelling techniques are still very evident and many researchers are investigating alternatives. section 4 of the paper therefore presents a selection of some of the work in progress on new models for CSR. section 5 then concludes.

## 2 BASIC MODELLING FRAMEWORK

### 2.1 PROBLEM FORMULATION

The statistical formulation of the speeech recognition problem assumes that speech can be represented by a sequence of acoustic vectors $Y = y_1 \ldots y_T$ and that this sequence encodes a sequence of words $W = w_1 .. w_K$. The specific form of the acoustic vectors is chosen so as to minimise the information lost in the encoding and to provide the best match with the distributional assumptions made by the subsequent acoustic modeling. In practice, a log spectral estimate[1] is typically computed every 10msecs and then a truncated cosine transformation is applied to smooth and partially decorrelate the feature elements. In addition, first order (delta) and second-order (delta-delta) regression coefficients are appended in a heuristic attempt to compensate for the conditional independence assumption made by the HMM-based acoustic models(see section 4.1). The result is a vector whose dimensionality is typically around 40 which has been partially but not fully decorrelated[6, Ch 5].

---

[1]the frequency spectrum is usually warped non-linearly to match the resolution of the human ear. The *mel* and *bark* scales are common approximations used for this warping

Recognition is then cast as a decoding problem in which we seek the word sequence $W^*$ such that[2]

$$W^* = \arg\max_W p(W|\mathbf{Y}) \qquad (1)$$

$$= \arg\max_W p(\mathbf{Y}|W)p(W) \qquad (2)$$

Here $p(\mathbf{Y}|W)$ is determined by an acoustic model and $p(W)$ is determined by a language model. The CSR problem is thus reduced to designing and estimating appropriate acoustic and language models, and finding an acceptable decoding strategy for determining $W^*$.

## 2.2 ACOUSTIC MODELLING

Since the vocabulary of possible words might be very large, the words in $W$ are decomposed into a sequence of basic sounds called *base phones* $Q$ of which there will be around 45 distinct types in English. To allow for the possibility of multiple pronunciations, the likelihood $p(\mathbf{Y}|W)$ can be computed over multiple pronunciations[3]

$$p(\mathbf{Y}|W) = \sum_Q p(\mathbf{Y}|Q)p(Q|W) \qquad (3)$$

where

$$p(Q|W) = \Pi_{k=1}^K p(Q_k|w_k) \qquad (4)$$

and where $p(Q_k|w_k)$ is the probability that word $w_k$ is pronounced by base phone sequence $Q_k = q_1^{(k)} q_2^{(k)} \ldots$. In practice, there will only be a very small number of possible $Q_k$ for each $w_k$ making the summation in eq 3 easily tractable.

Finally, each base phone $q$ is represented by a continuous density hidden Markov model (HMM) of the form illustrated in fig 1 with transition parameters $\{a_{ij}\}$ and output observation distributions $\{b_j()\}$. The latter are typically Gaussians and since the dimensionality of the acoustic vectors $\mathbf{y}_t$ is relatively high, the covariances are constrained to be diagonal.

Given the composite HMM $Q$ formed by concatenating all of the constituent base phones then the acoustic

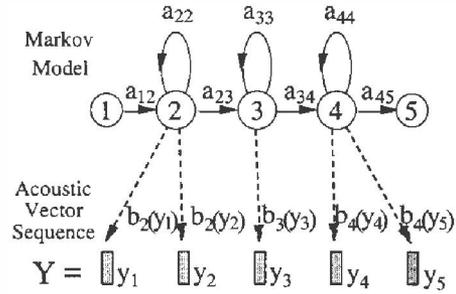

Figure 1: A simple HMM-based Phone Model

likelihood is given by

$$p(\mathbf{Y}|Q) = \sum_X p(X, \mathbf{Y}|Q) \qquad (5)$$

where $X = x(0)..x(T)$ is a state sequence through the composite model and

$$p(X, \mathbf{Y}|Q) = a_{x(0),x(1)} \Pi_{t=1}^T b_{x(t)}(\mathbf{y}_t) a_{x(t),x(t+1)} \qquad (6)$$

The acoustic model parameters $\{a_{ij}\}$ and $\{b_j()\}$ can be efficiently estimated from a corpus of training utterances using EM. For each utterance, the sequence of baseforms is found and the corresponding composite HMM constructed. A forward-backward alignment is then used to compute state occupation probabilities (the E-step), the means and variances are then maximised via simple weighted averages (the M-step)[6, Ch 7]. Note that in practice the majority of the model parameters are used to model the output distributions and the transition parameters have little effect on either the likelihood or the recognition accuracy.

The above approach to acoustic modelling is often referred to as the *beads-on-a-string* model, so-called because all speech utterances are represented by concatenating a sequence of precomputed phone models together.

## 2.3 LANGUAGE MODELLING

The probability of a word sequence $W = w_1..w_K$ is

$$p(W) = \Pi_{k=1}^K p(w_k|w_{k-1}, w_{k-2}, \ldots, w_1) \qquad (7)$$

For large vocabulary recognition, the conditioning word history in eq 7 is usually truncated to $n-1$ words to form an *N-Gram* language model

$$p(W) = \Pi_{k=1}^K p(w_k|w_{k-1}, w_{k-2}, \ldots, w_{k-n+1}) \qquad (8)$$

where $n$ is typically 2 or 3 and never more than 4. The n-gram probabilities are estimated from training texts by counting n-gram occurrences to form ML parameter estimates. The major difficulty of this method is data sparsity which is overcome by a combination of discounting and backing-off[7, 8].

---

[2]In this paper $p()$ is used to denote both a probability and a density, the context should indicate which is intended

[3]Recognizers often approximate this by a *max* operation

## 2.4 DECODING

Decoding is performed, conceptually at least, by compiling a network of all vocabulary words in parallel within a loop. Each word is represented by the subnetwork of HMM phone models corresponding to all of its allowable pronunciations. For the case of a bigram (N=2) language model, the transition probabilities between words are just the bigram probabilities. Once compiled, the whole *recognition network* can be used in a conventional Viterbi decoder to compute the most likely state sequence for an unknown input utterance. Trace back through this sequence then yields the most likely phone and word sequence.

In practice, however, decoding is more complex. Adequate performance requires at least a trigram (N=3) LM. This means that the recognition network must be expanded to allow all possible N-1 word histories to be maintained. Also, as will be explained in the next section, a small set of 40 or so base phones is inadequate to deal with the variability found in natural language and in practice, phone set sizes of 10000 or more are common with the choice of model being context-dependent. Thus, decoding is a complex search problem and decoder design continues to be a research issue[9].

Solutions to the decoding problem exploit sharing and pruning to limit the number of active hypotheses[10]. A standard scheme for reducing search costs uses multiple passes over the data. The output of each pass is a lattice of word sequence hypotheses rather than the single best sequence. This allows the output of one recognition pass to constrain the search in the next pass. Initial passes can use simple models when the search space is large and later passes can use more refined models when the search space is reduced. This multipass approach is also extremely convenient for research since it allows recognition experiments to be run on lattices rather than incurring the heavy computational cost of repeated full decodes[4].

## 3 THE STATE OF THE ART

The previous section has summarised the basic framework used by modern CSR systems. In doing so, a number of ill-found assumptions have been made such that a system built exactly as described would have rather poor performance. This section, examines these assumptions in more detail and describes technques used to mitigate them.

---

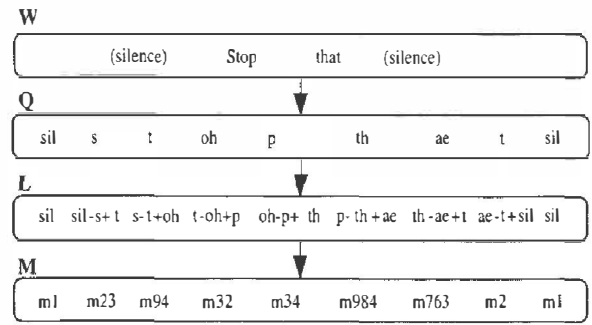

Figure 2: Context Dependent Phone Modelling

## 3.1 PHONOLOGICAL MODELLING

The simple approach described in section 2.2 of decomposing each vocabulary word into a sequence of base phones fails to capture the very large degree of context-dependent variation that exists in real speech. For example, the base form pronunciations for "mood" and "cool" would use the same vowel for "oo", yet in practice the realisations of "oo" in the two contexts are very different due to the influence of the preceding and following consonant. A simple way to mitigate this problem is to use a unique phone model for every possible context. To avoid the resulting data sparsity problems which would otherwise result, each of these *logical phones* $L$ can be mapped to a reduced set of shared *physical models* $M$[5]. Using this context-dependent phone decomposition, eq 3 conceptually becomes

$$p(\mathbf{Y}|W) = \sum_M p(\mathbf{Y}|M)\{\sum_L p(M|L)(\sum_Q p(L|Q)p(Q|W))\} \quad (9)$$

and the process is illustrated in fig 2 where the notation $x\text{-}y\text{+}z$ denotes the base phone y spoken in the context of a preceding x and a following z. Notice that the context-dependence spreads across word boundaries and this is essential for capturing many important phonological processes. For example, the [p] in "stop that" has its burst suppressed by the following consonant [th][6].

In eq 9, most current systems map the base phones $Q$ directly into the logical models $L$ and then cluster $L$ to form physical models $M$. Clustering typically operates at the state-level rather than the model level since it simplifies the tree computations and it allows a larger set of physical models to be robustly estimated. Fig 3 illustrates how the physical models are constructed by tying the states of the logical models.

---

[4]A recent alternative approach to decoding has been developed based on the systematic composition of weighted finite-state transducers[11].

[5]Here it is assumed that the context is determined by just the identity of the neighbouring phones. The resulting logical models are called *triphones*. However, further context can be included such as more distant neighbours, proximity to word boundaries, lexical stress, etc

[6]Note that this use of cross-word context dependent models greatly complicates the decoder.

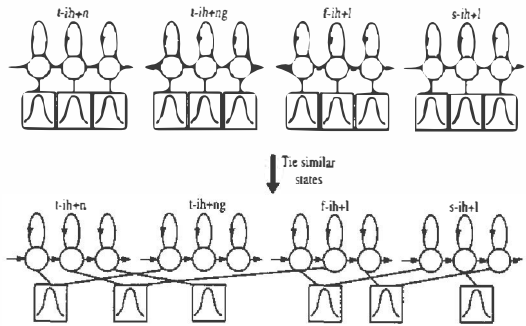

Figure 3: Formation of Tied-State Phone Models

The choice of which states to tie is made using decision trees[12]. Each state position[7] of each phone $q$ has a binary tree associated with it. Each node of the tree carries a question regarding the context. To cluster all states $i$ of phone $q$, all states $i$ of all of the logical models derived from $q$ are collected into a single pool at the root node of the tree. Depending on the answer at each node, the pool of states is successively split until all states have trickled down to leaf nodes. All states in each leaf node are then tied. The questions at each node are selected from a predetermined set to maximize the likelihood of the training data given the final set of state-tyings. Fig 4 illustrates this tree-based clustering. In the figure, the logical phones s-aw+n and t-aw+n will both be assigned to leaf node 3 and hence they will share the same central state of the representative physical model.[8]

The partitioning of states using phonetically-driven decision trees has several advantages. In particular, logical models which are required but were not seen at all in the training data can be easily synthesised. One disadvantage is that the partitioning can be rather coarse. This problem can be reduced using so-called *soft-tying*[13]. In this scheme, a post-processing stage groups each state with its one or two nearest neighbours and pools all of their Gaussians. Thus, the number of mixture weights in each state is increased whilst holding the total number of Gaussians in the system constant.

Decision-tree tied-state context dependent modelling schemes can handle most of the phonological variation found in carefully articulated speech (eg dictation). However, they fail to handle more radical phonolog-ical variations such as the processes which lead to "going to" being realised as [g ax n ax] ("gonna"). There have been a variety of attempts to handle this kind of problem within the *beads-on-a-string* framework by allowing more variation between the W and L layers eg by using decision trees to generate context dependent pronunciations[14, 15, 16]. However, none have been really successful. The essential problem is that expanding the set of possible pronunciations to give wider coverage of actual spoken forms, simultaneously increases confusibility with the result that recognition error rates improve little. Linguists have argued that this is a fundamental flaw in the *beads-on-a-string* framework[17].

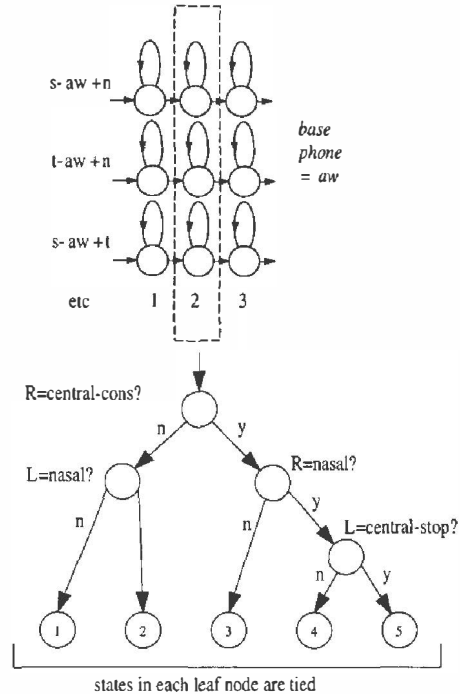

Figure 4: Decision Tree Clustering

## 3.2   IMPROVED DISTRIBUTIONAL MODELLING

In practice, speech feature vectors are not Gaussian and not uncorrelated. Furthermore, both intra-speaker and inter-speaker variability introduces modes in the data. Almost all CSR systems compensate for this by using a mixture of Gaussians to represent each state distribution, thus

$$b_j(\mathbf{y}_t) = \sum_{m=1}^{M} c_{jm} \mathcal{N}(\mathbf{y}_t; \boldsymbol{\mu}_m, \boldsymbol{\Sigma}_m) \qquad (10)$$

where the $c_{jm}$ are the mixture weights. Since, as explained in the previous section, there are typically a large number of these Gaussians[9], each component variance $\Sigma_\mathbf{m}$ is constrained to be diagonal.

---

[7]invariably each phone model has three states

[8]The total number of tied-states in a large vocabulary speaker independent system typical ranges between 5000 and 10000 states

The ability of mixtures of diagonal covariance Gaussians to model correlation structure is limited. One way to improve covariance modelling without incurring unacceptable training data and run-time computation requirements is to transform the input feature vector. A much studied approach to this is to use Linear Discriminant Analysis (LDA) and its variants[18, 19]. LDA can also be used to prune out *nuisance* dimensions from the feature vectors thereby reducing computation and potentially increasing classification performance. However, in practice, the gains achievable by LDA are variable, and more robust improvements have been obtained using Semi-tied Covariances[20]. This scheme decomposes each component covariance into the product $\hat{\Sigma} = \mathbf{H}^{-1}\Sigma(\mathbf{H}^{-1})^T$ where $\mathbf{H}$ is a transformation matrix and $\Sigma$ is diagonal. This has the property that

$$\log p(\mathbf{y}; \hat{\mu}_m, \hat{\Sigma}_m) = \log p(\mathbf{H}\mathbf{y}; \mu_m, \Sigma_m) + \log|\mathbf{H}| \quad (11)$$

Hence, if $\mathbf{H}$ is global, it can be applied as a transform to the speech data and the diagonal covariance Gaussians estimated and computed in the normal way on the transformed data. Although there is no direct EM-based reestimation for $\mathbf{H}$, a simple iterative scheme is available [20]. For large systems with many Gaussians, a single global transformation may be inadequate to capture the different correlation structures. In this case, a set of transformations $\mathbf{H}^{(\phi(m))}$ can be used, shared amongst all Gaussians via a partition function $\phi(m)$ chosen to maximise the training data likelihood. More recently, Gales has shown that $\mathbf{H}$ can be factored into streams to allow a larger set of transformations to be synthesised[21].

Another approach to dealing with inaccurate model assumptions is to use a discriminative objective function such as Maximum Mutual Information (MMI) as the basis for parameter estimation[22]. For MMI, the objective function is

$$\mathcal{F}(\lambda) = \log \frac{p_\lambda(\mathbf{Y}|M_W)P(W)}{\sum_{\hat{W}} p_\lambda(\mathbf{Y}|M_{\hat{W}})P(\hat{W})} \quad (12)$$

where $M_W$ is the composite model corresponding to the word sequence $W$ and $P(W)$ is the probability of this sequence. The summation in the denominator is taken over all word sequences $\hat{W}$ and it can be replaced by

$$p_\lambda(\mathbf{Y}|M_{\text{rec}}) = \sum_{\hat{W}} p_\lambda(\mathbf{Y}|M_{\hat{W}})P(\hat{W}) \quad (13)$$

where $M_{\text{rec}}$ encodes the full acoustic and language model used in recognition.

This objective function can be optimised using an extended form of Baum-Welch reestimation algorithm[23]. Furthermore, tractable approximations to the summations in the numerator and denominator terms of eq 12 can be computed from decoder generated lattices, where the numerator lattice is derived from a forced recognition of each training utterance and the denominator is derived from applying the full recognition model to each training utterance[24, 25].

### 3.3 NORMALISATION AND ADAPTATION

One fundamental assumption of the statistical framework is that the training data is representative of the unseen test data. In CSR this is rarely true. Not only do speakers vary greatly but also the background noise conditions and transducer channel characterisics are highly variable. The solution to this problem is to normalise the input data as much as possible and then use unsupervised adaptation to adjust the model parameters.

A very simple method of normalising for channel effects in *off-line* transcription applications is to subtract the cepstral mean and scale the variance of each feature element to unity. The region of speech used to compute the needed averages is constrained by the application but is typically the whole utterance or whole side of a conversation[26].

An effective form of speaker normalisation is to warp the frequency axis used in the front-end spectrum analysis in order to compensate for variations in vocal tract length[27, 26]. The optimal warping is found by searching for the scaling factor which maximises the log likelihood of the warped speech data.

Whereas normalisation seeks to make the input speech closer to the models, adaptation seeks to modify the models to make them a better fit to the speech. There are two main approaches to adaptation. Firstly, the model parameters can be treated as random variables and estimated using traditional Bayesian MAP techniques[28]. The problem with this approach is that only those parameters for which there is adaptation data are updated, also determining suitable priors can be difficult. The second and more widely used approach is to estimate transformations of the acoustic model parameters in a technique called Maximum Likelihood Linear Regression (MLLR)[29, 30, 31]. MLLR seeks to find an affine transform of the Gaussian means which maximises the likelihood of the adaptation data, i.e.

$$\hat{\mu}_m = \mathbf{A}\mu_m + \mathbf{b} = \mathbf{G}\eta_m \quad (14)$$

where $\mathbf{G} = [\mathbf{b}\ \mathbf{A}]$ and $\eta_m = \begin{bmatrix} 1\ \mu_m^T \end{bmatrix}^T$.

---

[9]typically 10 to 32 Gaussians per state, and typically 5000 to 10000 states in a system

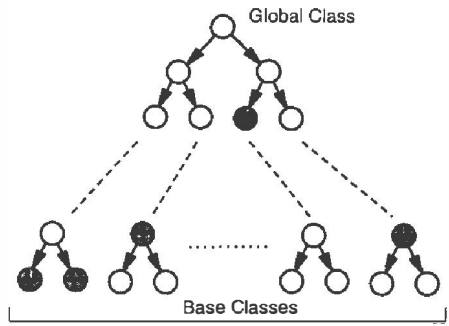

Figure 5: Regression Class Tree

The key to the power of this adaptation approach is that a single transformation **G** can be shared across a set of Gaussian mixture components. When the amount of adaptation data is limited, a single transform can be shared across all Gaussians in the system. As the amount of data increases, the HMM state components can be grouped into classes with each class having its own transform. As the amount of data increases further, the number of classes and therefore transforms increases correspondingly leading to better and better adaptation.

The number of transforms is usually determined automatically using a *regression class tree* as illustrated in fig 5. Each node represents a regression class i.e. a set of Gaussian components which will share a single transform. For a given set of adaptation data, the tree is descended and the most specific set of nodes is selected for which there is sufficient data (for example, the filled-in nodes in the figure).

In addition to mean adaptation, variance adaptation is also possible. There are several approaches to this. Firstly, a separate unconstrained transform **H** can be used where

$$\hat{\Sigma}_m^{-1} = \mathbf{C}_m \mathbf{H}^{-1} \mathbf{C}_m' \qquad (15)$$

and where $\mathbf{C}_m$ is the Choleski factor of $\Sigma_m^{-1}$. The advantage of this form is that **H** can be estimated using the standard covariance reestimation formula.

Secondly, instead of having a separate transform for the means and variances, a single *constrained* transform can be applied to both, i.e.

$$\hat{\mu}_m = \mathbf{A}\mu_m + \mathbf{b} \qquad (16)$$
$$\hat{\Sigma}_m = \mathbf{A}\Sigma_m \mathbf{A}' \qquad (17)$$

This has no closed-form solution but an iterative solution is possible [32]. A key advantage of this form of adaptation is that the likelihoods can be calculated as

$$\mathcal{L}(\mathbf{y}; \mu, \Sigma, \mathbf{A}, \mathbf{b}) = \mathcal{N}(\mathbf{Ay} + \mathbf{b}; \mu, \Sigma) + \log(|\mathbf{A}|)$$

This means that the transform can be applied to the data rather than the HMM parameters which may be more convenient for some applications (cf semi-tied covariances, section 3.2).

In addition to the above, there are many variants. For example, multiple transforms can be used for each class[33, 34] and alternative objective functions can be used[35, 36].

## 3.4 CONFUSION NETWORKS AND MODEL COMBINATION

As noted in section 2.1, the goal of the statistical formulation of the CSR problem is to find the most likely word sequence $W$. In the optimal case, this would yield the minimum sentence error rate whereas the pragmatic requirement is to minimise the word error rate (WER).

A minimum word error rate decoding can be achieved by estimating word posterior probabilities. This can be done by converting the decoder lattice output into a confusion matrix. For each arc in the lattice, a posterior probability can be computed using the forward-backward algorithm[10]. The lattice arcs can then be clustered to form a linear graph with the property that all parallel arcs form a confusion set and all paths through the graph pass through all nodes in the same order as in the lattice[37]. The minimum WER hypothesis is then found by selecting the most likely arc in each confusion set.

Confusion graphs can also be used to compute confidence scores[38]. The direct use of the word posteriors tends to overestimate the probability of correct recognition because the lattices from which they are derived only cover a fraction of the complete hypothesis space. Hence in practice some form of mapping is used (eg decision trees).

The above sections have described a range of techniques which can be used to improve the performance of a CSR system and these techniques can be combined in various ways. Empirically it is often found that even where different system combinations lead to similar performance, the errors made by each individual system are different. Hence a recent trend in transcription applications has been to combine the outputs of several decoders. A simple way of doing this is by voting[39]. However, combining the posteriors derived from confusion networks yields better performance[38].

---

[10]Note that in practice it necessary to scale-down the acoustic scores relative to the language model scores to avoid the posteriors being dominated by the most likely path.

## 3.5  LANGUAGE MODELLING

As described in section 2.3, the core of all current language models is the word n-gram. The n-gram captures local syntactic and semantic dependencies and for many languages this is sufficient to cover a large fraction of the useful constraints. Its primary weakness is that it is inevitably undertrained[11] and although smoothing and backing-off procedures can mitigate this effect, data sparsity is always a problem.

One way to reduce the effects of data sparsity is to interpolate the word n-gram with a class-based language model[40]. A class-based LM maps all words into a relatively small number of classes for which n-grams can be robustly estimated, i.e.

$$p(w_k) = p(c_k|c_{k-1}, \ldots, c_{k-n+1})p(w_k|c_k) \qquad (18)$$

where it is assumed that every word maps to a unique class. The classes themselves $\{c_i\}$ can be determined automatically using an ML-based clustering algorithm[41]. Empirically it is found that 250 to 1000 classes give the best performance.

## 4  WORK IN PROGRESS

As the previous section has shown, there are many techniques which can compensate for the ill-found assumptions of the HMM-based recognition framework. Furthermore, when these techniques are carefully integrated, the result in terms of performance can be very competitive. For example, the first pass of the HTK large vocabulary recogniser consists of simple tied-state cross-word triphones trained using maximum likelihood and a trigram language model. When applied to the March 2000 Hub5[12] evaluation data, the word error rate was 38.6%. However, with the addition of VTLN and MLLR adaptation, quinphone acoustic models, soft-tied states, semi-tied covariances, MMI training, 4-gram language model interpolated with trigram class model, confusion graph scoring, and fourway model combination, the error rate was reduced to 25.4%, i.e. a relative reduction of 34%[42]. This is a very demanding task, and no other system or architecture has managed to come close to this level of performance.

Despite this impressive performance there are three key assumptions in current CSR systems which continue to cause concern:

• the *frame-independence* assumption whereby each

successive speech feature vector is assumed independent

• the *beads-on-a-string model combination* whereby phone-based HMMs are concatenated in sequence to form words and sentences

• the *n-gram language modelling* which prevents modelling of long-range dependencies

It is very easy to argue the case for each of these assumptions being a major limitation on the performance of CSR systems. Nevertheless, they are all extremely resistent to improvement. The next few sections will provide some pointers to recent and current work which attempts to improve on these assumptions.

### 4.1  SEGMENT MODELS

Attempts to weaken the frame-independence assumption exploit the concept of a *segment model* whereby speech features are segment[13] rather than frame-based (see [43] for a review). In this framework, a segment of frames $\mathbf{y}_1 \ldots \mathbf{y}_l$ representing phone $q$ is modelled as

$$p(\mathbf{y}_1 \ldots \mathbf{y}_l, l|q) = p(\mathbf{y}_1 \ldots \mathbf{y}_l|l, q)p(l|q) \qquad (19)$$

where the first term on the right is the observation distribution and the second term is the segment duration distribution[14].

The characteristics of a segment model are determined by the form of the observation distribution. The simplest form maps the segment into fixed regions $r_i$ and associates a specific distribution with each region

$$p(\mathbf{y}_1 .. \mathbf{y}_l|l, q) = \Pi_{t=1}^l p(\mathbf{y}_i|q, r_i) \qquad (20)$$

where the mapping from frames to regions can be chosen to maximise the likelihood.

Rather than associating a distribution with each region, distributions can be parameterised to define trajectories $f_a(t)$ through the segment. The probability of a segment is then

$$p(\mathbf{y}_1 .. \mathbf{y}_l|l, q) = \int_a \Pi_{t=1}^l p(\mathbf{y}_t|f_a(t))da \qquad (21)$$

Typically the trajectory defines the evolution of the intra-segment mean and this can be fixed as in [44] or linear as in [45]. Rather than integrating over all possible trajectories, it is usually more computationally convenient to find the most likely trajectory.

---

[11]Consider that for $n = 3$, a 50k word vocabulary requires more than $10^{14}$ trigrams to be estimated.

[12]i.e. "Switchboard" and "Call Home", conversational speech over the telephone

[13]In this context, a segment is equivalent to a phone in the terminology of earlier sections.

[14]Decoding requires that the durational term be integrated out, hence the computational complexity of segment models is significantly higher than conventional HMMs

Perhaps the most appealing form is the linear dynamical system since this can be related to the dynamics of the underlying human production system

$$\mathbf{x}_{t+1} = F_t\mathbf{x}_t + \mathbf{w}_t; \quad \mathbf{y}_t = H_t\mathbf{x}_t + \mathbf{v}_t \qquad (22)$$

where $\mathbf{x}_t$ is a state vector and $\mathbf{u}_t$ and $\mathbf{v}_t$ are Gaussian processes. This model, well-known, in control applications was introduced into speech by Digalakis[46]. It is also receiving renewed interest in the machine learning community[47]. In speech, parameters defining $\mathbf{u}_t$ and $\mathbf{v}_t$ are associated with each region of each segment. The probability of each segment is then computed via the innovation sequence $\{\mathbf{e}_t\}$

$$p(\mathbf{y}_1..\mathbf{y}_l|l, q) = \Pi_{t=1}^l p(\mathbf{e}_t|q, r_t) \qquad (23)$$

In effect, $\mathbf{x}_t$ defines a hidden trajectory over the segment from which the observations $\mathbf{y}_t$ are derived. This has led to variety of derivatives some of which use articulator-based models for the trajectory $\mathbf{x}_t$ and nonlinear mappings from $\mathbf{x}_t$ into $\mathbf{y}_t$[48, 49].

Although much good work has been done on segment models, the results so far have been disappointing. One reason for this might be that more precise modelling of segments simply exacerbates the errors caused by the *beads-on-a-string* assumption.

## 4.2   ASYNCHRONOUS PARALLEL MODELS

The *beads-on-a-string* model combined with context-dependent phone models can handle the variations found in carefully articulated speech but it fails when significant phonological variation occurs as in everyday spontaneous speech. Many phonological processes are more naturally expressed in terms of a hierarchy of parallel feature streams [17]. For example, when nasality from a nasal consonant colours a neighbouring vowel, this would be modelled by differences in the timing of feature changes rather than as a substitution of one allophone for another. In this model, pronunciation variability is due to asynchrony between feature changes, and although the feature tiers are asynchronous they are nevertheless coupled. At the signal level, similar effects can be observed. Formants do not move synchronously and empirically there is support for separate modelling of frequency bands.

Observations such as these have motivated study on good ways to model parallel asynchronous processes. These range from very loosely coupled Multi-band models where streams are modelled independently and synchronised at major (eg syllable) boundaries[50], to very tightly coupled models where multiple observation distributions share the same underlying Markov chain[6].

More recently, these parallel stream models have been generalised to allow the required degree of coupling to be learnt via the framework of *factorial HMMs*. A factorial HMM consists of $L$ individual Markov chains or streams evolving in parallel and $K$ observation distributions. At any time, metastate $I$ of the model is determined by the state of each individual stream $I = (i^1, \ldots, i^L)$ and the observation $\mathbf{Y}_t$ is composed of $K$ components $\mathbf{Y}_t = (\mathbf{y}_t^1, \ldots, \mathbf{y}_t^K)$. Metastates are conditionally independent given the previous metastate, and observations are conditionally independent given the current metastate, i.e.

$$p(J|I) = \Pi_{l=1}^L p(j^l|I) \qquad (24)$$

$$p(\mathbf{Y}_t|J) = \Pi_{k=1}^K p(\mathbf{y}_t^k|J) \qquad (25)$$

This structure provides a flexible framework for modelling asynchronous processes but the state spaces need to be constrained to be tractable. Current schemes for doing this include a mixed-memory approximation[51] and parameter tying[52]. Training and decoding using these models is potentially very expensive but a simple sub-optimal *Chain Viterbi* scheme where each stream is aligned in turn keeping the other streams fixed has been shown empirically to be effective[53].

The practical application of parallel asynchronous HMM's is still in its infancy but there are several existing research efforts that might benefit from it including those working on hidden articulator models[54, 55] and those working on discriminative feature models[56]. Also, Hinton has shown how the framework of parallel HMMs can be extended further[57].

## 4.3   LANGUAGE MODELLING

As noted earlier, the main limitation of n-gram language models is that they can only model local dependencies. Thus, for example, in "the warden locked the cell door", "the warden" is a good predictor for "locked" whereas in "the warden with a limp locked the cell door", "a limp" is a poor predictor of "locked".

Early attempts to extend statistical LMs to include longer range dependencies mostly focussed on *trigger* models whereby predictor words are counted if they lay anywhere within the history. Triggers can be conveniently combined with conventional n-grams using the Maximum Entropy framework (ME) which leads to solutions with the following exponential form[58]

$$p(w_k|h_k) = \frac{1}{Z}\Pi_i e^{\lambda_i f_i(w_k, h_k)} \qquad (26)$$

where $h_k$ denotes the history $w_{k-1}\ldots$ of word $w_k$. The functions $f_i(w_k, h_k)$ are constraint functions, and the $\lambda_i$ are chosen to ensure that the expected values of the $f_i$ match to $p(w_k|h_k)$ matches the known marginals. For example, a bigram constraint would be $f_{w_1,w_2}(w,h) = 1$ iff $w = w_2$ and $h$ ends in $w_1$, and the required marginal constraint would be given by the count $\#(w_1, w_2)/\#(w_1)$ computed over the training data. Within this framework, triggers are easily incorporated as constraints of the form $f_{w_1,w_2}(w,h) = 1$ iff $w = w_2$ and $w_1 \in h$.

More explicit approaches to exploiting syntactic and semantic models use probabalistic parsers to uncover *head words* which can then be used as predictors[59]. Using ME, these can be combined with conventional n-gram constraints[60]. This work is especially interesting since it models longer range dependencies in a more principled way than triggers. In the longer term, the growing synergy between the statistical approaches to speech and computational linguistics should pay dividends in this area.

## 5 CONCLUSIONS

This paper has reviewed the statistical framework used to build continuous speech recognition systems and briefly described the most important refinements needed to endow such a system with state-of-the-art performance. The three key assumptions underlying current approaches are (a) frame-independence; (b) beads-on-a-string model combination; and (c) n-gram language modelling. The latter part of the paper has described work in progess which aims to improve upon these assumptions. So far, significant gains in performance resulting from this newer work have been sadly lacking. However, as argued in [61], it is important that such work continues even if it does mean increasing error rates in the short term. Modern start-of-the-art systems are impressive and they will improve further. Nevertheless, the three fundamental assumptions on which they are based must surely mean that they are climbing to a local maximum, and somewhere there are better solutions . . .